# EXACT: TOWARDS A PLATFORM FOR EMPIRICALLY BENCHMARKING MACHINE LEARNING MODEL EXPLANATION METHODS

*Benedict Clark* [a,*], *Rick Wilming* [b], *Artur Dox* [a,b],
*Paul Eschenbach* [b], *Sami Hached* [b], *Daniel Jin Wodke* [b], *Michias Taye Zewdie* [b], *Uladzislau Bruila* [b],
*Marta Oliveira* [a], *Hjalmar Schulz* [b,c], *Luca Matteo Cornils* [b], *Danny Panknin* [a], *Ahcène Boubekki* [a], *Stefan Haufe* [a,b,c]

[a] Physikalisch-Technische Bundesanstalt, Abbestrasse 2-12, 10587 Berlin, Germany, email address: stefan.haufe@ptb.de
[b] Technische Universität Berlin, Str. des 17. Juni 135, 10623 Berlin, Germany, email address: haufe@tu-berlin.de
[c] Charité – Universitätsmedizin Berlin, Charitéplatz 1, 10117 Berlin, Germany, email address: haufe@tu-berlin.de
* Corresponding author

*Abstract* - The evolving landscape of explainable artificial intelligence (XAI) aims to improve the interpretability of intricate machine learning (ML) models, yet faces challenges in formalisation and empirical validation, being an inherently unsupervised process. In this paper, we bring together various benchmark datasets and novel performance metrics in an initial benchmarking platform, the Explainable AI Comparison Toolkit (EXACT), providing a standardised foundation for evaluating XAI methods. Our datasets incorporate ground truth explanations for class-conditional features, and leveraging novel quantitative metrics, this platform assesses the performance of post-hoc XAI methods in the quality of the explanations they produce. Our recent findings have highlighted the limitations of popular XAI methods, as they often struggle to surpass random baselines, attributing significance to irrelevant features. Moreover, we show the variability in explanations derived from different equally performing model architectures. This initial benchmarking platform therefore aims to allow XAI researchers to test and assure the high quality of their newly developed methods.

## 1. INTRODUCTION

Research in the field of Explainable AI (XAI) aims to 'explain' the decisions of complicated Machine Learning (ML) models, with authors aiming to deploy their methods to high stakes domains such as medicine and law [1–3]. In recent years, a plethora of XAI methods have been developed to achieve this goal. In the past, methods such as SHAP [4] or LIME [5] have emerged as popular choices to assess the quality of ML models. In addition, the quality and robustness of such methods have been assessed by various supporting evaluation studies, already highlighting weaknesses of such methods. However, it remains unclear what to conclude from the output of XAI methods in general, since there is a lack of a formal problem definition of explainability. Being an inherently unsupervised task, formalization and empirical validation of the quality of explanations produced is difficult and limits their potential use for quality-control and transparency purposes. As such, current research often tends to rely on subjective evaluation of methods, for example through user studies on which of two given explanations appear better qualitatively [6] as well as evaluation of secondary properties of explanation methods [5]. Often these evaluation studies do not contain a formal definition of an explanation, but this is important to be able to interpret explanations correctly and understand their limits. When faced with so-called suppressor variables, in the context of explanations, high importance may be attributed to these types of variables although they lack any statistical relation to the prediction target [7]. The inclusion of suppressors may allow a model to remove unwanted noise, which can lead to improved prediction quality. While it is clear suppressors can be useful for a model, it has been shown empirically that many of the most popular XAI methods also highlight suppressor features as important [8], which may lead to misinterpretations for the user. In the context of image or photography data, suppressor variables could be encapsulated by background pixels with lighting information. The model could normalise the brightness in the image to achieve a better prediction, and an explanation method may then highlight such background pixels as important. Thus, accounting for potential emergent features carrying properties as they can be found in suppressor variables, is essential for an objective benchmark assessing the correctness of machine learning explanations.

Recent benchmarks have begun to tackle this across a variety of individual linear and non-linear benchmarks in synthetic image domains as well as medical image classification [9–11]. Experimental evidence backs up the theoretical claims that XAI methods are hindered in the quality of the explanations they produce by highlighting suppressor variables as important and can perform worse quantitatively than simple edge detection methods.

Most aforementioned benchmarks are also inherently limited to the list of methods chosen by the authors, so it is difficult to draw comparisons across two methods evaluated in separate benchmarks. Further benchmarks are then also limited by the rate of publishing new research and will always remain a static (albeit perhaps shifting over time) set of XAI methods.

In this paper, we aim to bring together several image benchmarks across different synthetic and semi-synthetic domains [9–11] to a unified challenge-style benchmark platform. Existing work already objectively benchmarks over 14 XAI methods and several performance baselines, and we lay down the foundation for authors of new methods to submit their work for automatic benchmarking. Specifically, we propose a prototype of a docker-based web application where users can submit the code of their explanation method in a standardised format to various 'challenges', composed of ground truth XAI benchmarks across several domains, where they will receive automatic scoring and placement on a leaderboard to compare all evaluated methods.

As a result, this work contributes to general standardization efforts allowing for objective assessments of existing and new XAI methods and therefore advancing the quality assurance of machine learning systems.

## 2. METHODS, DATASETS, AND METRICS

The general workflow of applying post-hoc Explainable AI techniques is as follows: given a dataset, we train a machine learning model using the training (and validation) split of the data. Taking the trained model and test data (either as individual samples or a batch) as the inputs to the XAI method, we receive output explanations of the same dimensionality as the input data, aimed to correspond to the 'importance' of each pixel towards the trained model's prediction output. This application of XAI methods to data after model training and (mostly) independent of model choice is why such methods are dubbed 'post-hoc'. Finally, we apply novel performance metrics to compare produced explanations and the ground truth for the given sample, giving the explanation performance of the method.

Prior work has studied the empirical performance of up to 16 existing XAI methods from the Captum [12] and iNNvestigate [13] packages [9–11]. These packages make the usage of XAI quick, and results from these prior studies form the backbone of performance comparison to future methods.

Based on prior published studies and planned work, we present the following benchmark datasets with explicitly defined ground truths for explanations. Presently, the prototype has been tested on the XAI-TRIS benchmarks [10], however all stated datasets will be available for users by the time of publication. We initially plan to integrate three image classification-based benchmarks – two synthetic with varying underlying statistics [9,10], and one semi-synthetic Magnetic Resonance-based benchmark [11]. We also are exploring other modalities such as natural language processing and tabular benchmarks, and while these are not yet published benchmarks in their own right, we outline the general concepts and how they would be integrated into the benchmark platform in the future.

### 2.1 XAI-TRIS

Leveraging prior work [8], we make use of tetrominoes [14] to form the main benchmark datasets used in the prototype platform. Tetrominoes are geometric shapes composed of four blocks, made famous by the popular game Tetris [15]. The XAI-TRIS datasets are composed of one linear and three non-linear binary image classification problems based on such tetrominoes, integrating these shapes with three background noise types (white, correlated, and imagenet). In total, these form 12 binary classification problems – each able to be considered its own dataset – of varying difficulty. Primary analysis is done with images of size $64 \times 64$-px, however initial analysis and testing was on scaled down $8 \times 8$-px equivalent scenarios. One benefit of the dataset is the ability to arbitrarily scale the dimensionality of the images and tetrominoes up and down to suit the user's needs.

Data is generated by combining tetromino shapes (signal $a$ determining the class-conditional distributions) with some noisy background $\eta$ according to two processes, one additive and multiplicative. By carefully placing tetromino patterns in each sample we define our binary classification problems, for example classifying a 'T'-tetromino in the top left of the sample versus an 'L'-tetromino in the bottom right of another. By adding these patterns together with the background component, we create a linear problem. Alternatively, multiplying signal tetromino patterns with the background component turns this into a non-linear problem. This forms the basis of the four main classification scenarios detailed later in this section, but first we define the additive process

$$x = \alpha\big(R \circ (H \circ a)\big) + (1 - \alpha)(G \circ \eta) \qquad (1)$$

as the combination of signal $a$ and white Gaussian background noise $\eta \sim N(\mathbf{0}, \mathbf{I})$, forming sample $x$. For the $64 \times 64$ data, the signal component (containing the tetromino patterns defining the classification problem) undergoes a 2D gaussian smoothing operation $H$ to smooth the integration of the pattern's edges into the background. The operation $R$ is a random spatial rigid body transformation, either the identity operation for three of the classification scenarios, or a random translation and 90-degree rotation for one of the scenarios. Each classification problem also has a second possible background type, where the Gaussian spatial smoothing filter $G$ is applied to $\eta$ to create a smoothed background with correlations between features of $\eta$. The third background type is that of replacing $G \circ \eta$ with samples from the ImageNet database [16]. Each sample is cropped and scaled to $64 \times 64$-px, preserving the original aspect ratio. This background type allows for a better relation of results to 'real world' scenarios due to the wide variety of different and complex structures in the background of generated samples. Transformed signal and noise components are normalised by their Frobenius norms, and the weighted sum of signal and background components is calculated, with the scalar parameter $\alpha \in [0,1]$ determining the signal-to-noise ratio (SNR).

The multiplicative generation process

$$x = \big(\mathbf{1} - \alpha\big(R \circ (H \circ a)\big)\big)(G \circ \eta) \qquad (2)$$

follows the same general structure and nomenclature, but instead presents a multiplication between the transformed signal and background components. This multiplication allows for non-linearity to be encapsulated in the data generation process.

All data generated according to either process are scaled to the range $[-1,1]$, such that $x \leftarrow x / \max|x|$, where $\max|x|$ is the maximum absolute feature value of the dataset.

*Aside: suppressor variables*

It is important to note that the correlations between features of $G \circ \eta$ in the case of using correlated noise induces the presence of suppressor variables. Here, a suppressor would be a pixel not part of the foreground $R \circ (H \circ a)$ that, however, correlates with a pixel of the foreground due to the smoothing operator $G$. Based on past work on the characteristics of suppressors [7,17,18] as well as other recent works benchmarking XAI methods in the presence of suppressor variables [8,9], the XAI-TRIS benchmarks also showed the susceptibility of XAI methods to highlighting such suppressors as important variables. This susceptibility has been shown in this study as well as the other aforementioned benchmarks to lead to drops in explanation performance.

*Scenarios: Linear (LIN) and Multiplicative (MULT)*

For the linear case, the additive generation model of Eq. (1) is used, and for the multiplicative case, the multiplicative generation model of Eq. (2) is used instead. In both, signal patterns are defined as a 'T'-shaped tetromino pattern $a^T$ near the top left corner if $y = 0$ and an 'L'-shaped tetromino pattern $a^L$ near the bottom-right corner if $y = 0$, leading to the binary classification problem. Each pattern is encoded as a mask such that $a_{i,j} = 1$ for each pixel in the tetromino pattern, positioned at the i-th row and j-th column, and zero otherwise. Each of the four 'blocks' of the tetromino is $8 \times 8$-px in size.

*Scenarios: Translations and rotations (RIGID)*

In this scenario, the 'T'- and 'L'-tetromino patterns defining each class are no longer in fixed positions but are transformed by the rigid body transformation $R$, corresponding to a random translation as well as a random rotation by multiples of 90 degrees. This is constrained such that the entire tetromino is contained within the image. In contrast to the other scenarios, a 4-pixel thick tetromino here to enable a larger set of transformations, and thus increase the complexity of the problem. This is an additive manipulation in accordance with Eq. (1).

*Scenarios: Translations and rotations (RIGID)*

The final scenario is an additive XOR problem, where we use both tetromino variants $a^T, a^L$ in every sample. Class membership is defined such that members of the class where $y = 0$, combine both tetrominoes with the background of the image either positively or negatively, such that $a^{XOR++} = a^T + a^L$ and $a^{XOR--} = -a^T - a^L$. Members of the opposing class, where $y = 1$, imprint one shape positively, and the other negatively, such that $a^{XOR+-} = a^T - a^L$ and $a^{XOR-+} = -a^T + a^L$. Each of the four XOR cases are equally represented across the dataset. Tetromino blocks are also 8-px thick here.

The ground truth feature set of important pixels to be used as the 'ideal' explanation is given by any non-zero pixel in $R \circ (H \circ a)$. In the LIN and MULT cases, as the tetrominoes for each class are in fixed positions, the combined ground truth set of tetromino patterns for both classes are used. This is because the presence of one tetromino at one location is just as informative as the absence of the other tetromino at the other location. This is equivalent to the operationalisation of feature importance seen in [9], specified in Section 2.3.

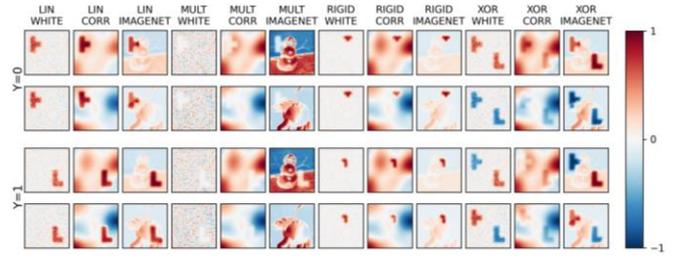

Figure 1. Example of data for each XAI-TRIS scenario, showing two samples for each class. Figure taken from [10].

## 2.2 Brain MRI dataset with superimposed artificial lesions

The next dataset is that of a semi-synthetic Magnetic Resonance Imaging (MRI) dataset [11] designed to mirror a realistic classification task. The underlying data is 2-dimensional T1-weighted axial MRI-slices sourced from the Human Connectome Project (HCP, [19]). Specifically, the healthy brain HCP dataset is used, composed of 3D MRI data sourced from 1007 healthy adults between the ages of 22 and 37 years old. 3D MRI slices are pre-processed with the FSL [20] and FreeSurfer [21] tools, and defaced as seen in [22]. For this study, slices with less than 55% black pixels are used, with $260 \times 311$-px images being padded vertically with zeros and cropped horizontally to yield $270 \times 270$ images.

Each resulting 2D axial slice provides the background for a random number of artificial 'lesions' to be overlaid on, with a binary classification between regular and irregularly-shaped lesions forming the task here. Lesions are generated from a $256 \times 256$-px pixel noise image, smoothed with a Gaussian filter of radius 2 pixels. The smoothed image is binarized by the Otsu method [23], and the opening and erosion morphological operations are applied. A second erosion is applied to create more irregular shapes. Connected components of the resulting noise map are identified to serve as potential lesions, and are selected based on the compactness of their shape. Considering the isoperimetric inequality on a plane $A \leq p^2/4\pi$, where $A$ is the area of a particular lesion and $p$ is its perimeter. Compactness is obtained by comparing the shape of the given lesion to a circle of equal perimeter, where a larger compactness leads to a rounder shape.

Three to five lesions of the same type (either regular or irregular) are assigned to random positions in a binary mask, imposed into a given sample by a pixel-wise multiplication with the background, preventing overlapping between lesions and ensuring presence within brain matter. Lesions are composed of intensity values $L_{i,j} \in [0, w]$ where $w$ is a constant controlling the SNR, with higher $w$ values leading to whiter and more visible lesions. In this study, the intensity value $w = 0.5$ is used. Resulting lesions appear like white matter hyperintensities, considered important biomarkers of the ageing brain and ageing-related neurodegenerative disorders [24,25].

This study also invokes the presence of suppressor variables, here being background pixels outside of the lesion which can provide the model with information on how to remove underlying brain content from the lesion area to improve the model's classification.

The ground truth feature set of important pixels are summarised as any non-zero value of the lesion mask which is subsequently multiplied with the underlying healthy brain data.

The underlying problem studied with this dataset initially [11] is the impact of transfer learning on the quality of explanations produced, where experimental results show that a deeper pre-training, as well as pre-training on the same underlying corpus, results in higher quality explanations.

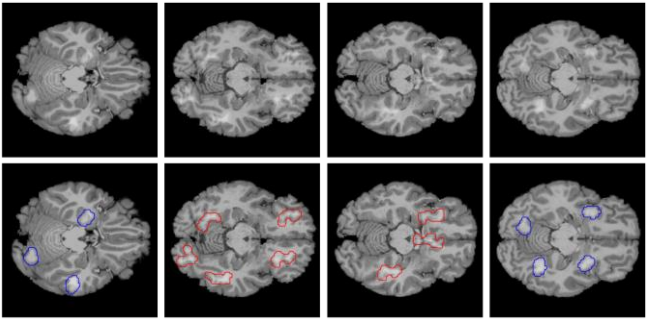

Figure 2. Example of axial MRI slices from the Human Connectome Project (HCP, [19]) healthy brain dataset with artificial lesions imposed (top), with the position of each lesion contoured in blue and red corresponding to each class (round versus irregular, bottom). Figure taken from [11].

### 2.3 Linear data with a salient distractor pattern

Both previous datasets incorporate suppressor variables in the overlap of background pixels with truly important tetrominoes and lesions, however these suppressors are not exactly specified and quantified. Controlling and measuring the exact suppression present in data is also a desirable property to benchmark XAI methods in a similar manner. The work of [9] takes a linear scenario where a signal activation pattern composed of two blobs in the left side of an $8 \times 8$-px image overlaps additively in the top-left with a distractor pattern of two equally-sized blobs in the top of the image. The activity within the distractor component is uncorrelated with the classification target, making it an instance of a particularly salient suppressor.

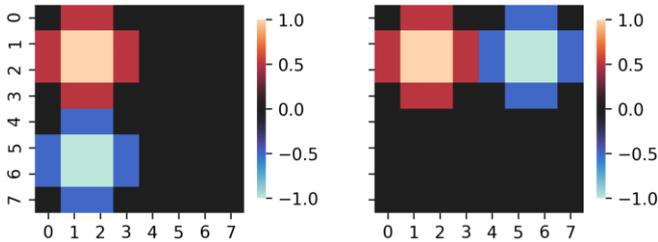

Figure 3. Signal activation pattern (left) compared to the distractor pattern (right) for the 'two blob' data, where the data is combined linearly with Gaussian noise added to form each sample. The overlap between the two red blobs in the top left result in the distractor pattern acting as suppressor variables. Figure taken from [9].

### 2.4 Natural language and tabular benchmarks

It is important to consider other modalities other than image data. Incorporating similar design principles of explicitly generated ground truth explanations, we plan to extend the available set of datasets to the realms of natural language processing and tabular datasets. The work of [26] considers synthetic tabular data with explicitly defined ground truth explanations, however do not focus on the issue of 'correctness' in explanations through the potential presence of suppressor variables, and evaluate XAI methods through subjective performance metrics such as faithfulness. These issues are what will be targeted in our future benchmark datasets.

### 2.5 Explanation performance metrics

With the datasets defined, we also need appropriate metrics to evaluate explanations produced. In the EXACT platform, user-submitted XAI method code will be evaluated on the test datasets specified above, with the following metrics calculated to compare explanations produced to the ground truth feature sets given by each benchmark dataset. Again, leaning on prior work, we focus the prototype on three core metrics – Precision, Earth Mover's Distance (EMD) and Importance Mass Accuracy (IMA).

#### Precision

Taking k as the number of non-zero pixels in the ground truth feature set, we define Precision as the fraction of the top-k features of the explanation which overlap with the features of the truly important ground truth, divided by k.
A score of 1 means that every one of the top-k features of the explanation are also the k non-zero features of the ground truth set.

#### Earth mover's distance (EMD)

Following [10], we define a performance metric based on the optimal cost of transforming one distribution to another. Applying this principle to the cost required to transform a continuous-valued explanation $s$ into the ground truth set of truly important features $F^+$, we normalise both inputs to have the same mass. Following the algorithm proposed in [27] and the implementation of [28], the Euclidean distance between pixels is used as the ground metric and the optimal transport cost $OT(s, F^+)$ is calculated for a given sample. The normalised performance score is defined as

$$EMD = 1 - \frac{OT(s, F^+)}{\delta_{max}}, \quad (3)$$

where $\delta_{max}$ is the maximum Euclidean distance between any two pixels in the sample, i.e. the distance of transporting mass from one corner of the image to the opposite corner.
A score of 1 means that the explanation perfectly and evenly covers the ground truth feature set, with no transport required to transport one to the other.

Note that the ground truth of important features is completely based on the data generation process. It is conceivable, though, that a model uses only a subset of these for its prediction, which must be considered equally correct.
The above performance metrics do not fully achieve invariance in that respect. However, both are designed to de-emphasize the impact of false-negative omissions of features in the ground truth on performance, while emphasizing the impact of false-positive attributions of importance to pixels not contained in the ground truth.

#### Importance Mass Accuracy (IMA)

To combat this, we define a third metric as the sum of importance attributed to ground truth features over the total attribution in the explanation, as seen in [10,29]. This metric overcomes the 'subset problem' proposed above whilst using the full explanation compared to Precision, which just looks at the 'top-k' features of the explanation.

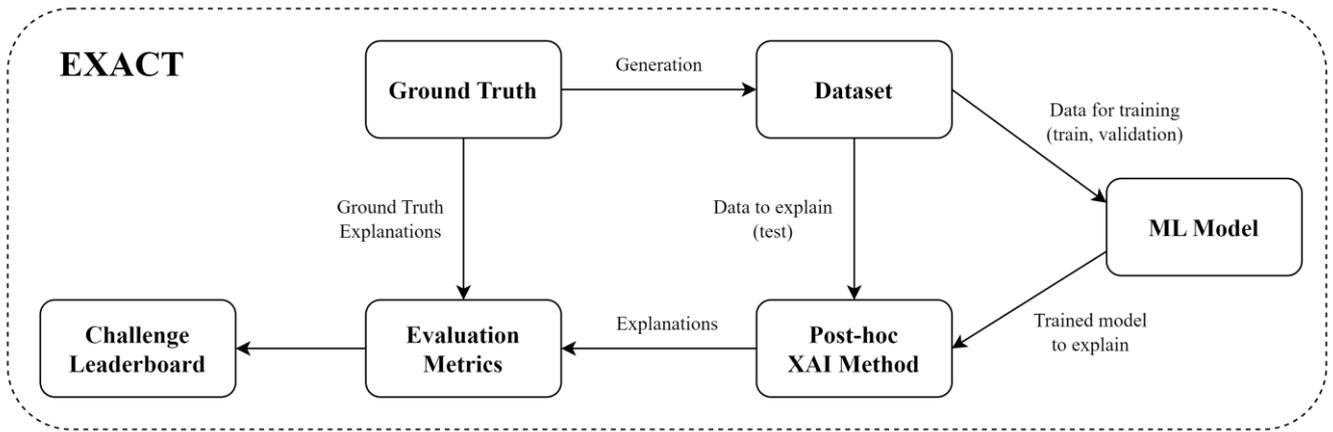

Figure 4. The process of evaluating an XAI method using the EXACT benchmarking platform. Classification datasets are generated through an explicitly known ground truth controlling the class-conditional distribution, which serves as the ground truth for explanations. Given an ML model trained on the given data, the user's XAI method takes test data and the model as input, producing explanations. These explanations are passed to the novel performance metrics, which use the given ground truth as a basis for comparison. Performance scores are stored in a leaderboard to compare the most and least performant XAI methods.

IMA is also a direct measure of false positive attribution, where a score of 1 means that the explanation perfectly highlights only the truly important features as important.

## 3. EXACT: THE EXPLAINABLE AI COMPARISON TOOLKIT

EXACT provides researchers with the opportunity to evaluate the correctness of explanations produced by XAI methods. Practitioners have the ability to upload the code for their newly developed explanation method to the platform and have its explanation performance assessed in comparison to many of the most popular current methods in the field.

### 3.1 Evaluating an XAI method

The process of evaluating an XAI method using EXACT can be seen in Figure 4, with a detailed API diagram shown in Figure 5. We can break this down into four key areas:

*Datasets*

Datasets in EXACT are as specified in Section 2, generated through explicitly known ground truth signals which serve as the ground truth for explanations. As can be seen in Figure 2, data is used both to train the given machine learning models, and also to serve as an input to the user's XAI method.

On the platform, each dataset is considered its own 'challenge' that the user can submit their XAI method to. Generated data is split, and the training and validation data is provided to the user to download on the challenge page. The test data and ground truth labels are reserved for use by the platform internally, to ensure integrity of the submitted code.

*Models*

Once trained, a machine learning model is provided as the second input to the XAI method. For the focus of evaluating false-positive attribution of feature importance by XAI methods, it is required that the model should ideally learn only patterns from the class-conditional ground truth distribution of our datasets. This allows direct comparison to ground truth pattern masks as the measure of explanation performance. In practice, we see that machine learning models make use of other non-dependent features (such as suppressors) by evaluating the explanations they produce [8-11].

In each challenge of the EXACT platform, the user can download not only the reference data, but also the pre-trained machine learning model to locally test their XAI method code. Providing a pre-trained model allows for consistency in evaluation, as model prediction accuracy has been shown to affect explanation performance [29]. We make use of the models used in the prior studies on the associated datasets. For example, challenges for the XAI-TRIS dataset outlined in Section 2.1 make use of a Linear Logistic Regression model, a Multi-Layer Perceptron, and a Convolutional Neural Network [10]. These all have different properties and capabilities, and are specific to the challenge dataset.

While we currently are focusing on benchmarking post-hoc XAI methods, a future iteration can open the possibility of uploading the user's own model for the case of models offering 'intrinsic' explainability.

*XAI Methods*

Given a trained machine learning model and the test data to be explained as an input, an XAI method produces 'explanations' of which input features are considered important by the model. We expect that an explanation produced should focus on highlighting the ground truth features of the dataset as important. False positive attribution of importance to other features, such as suppressor features, can lead to misinterpretation in practice.

EXACT currently makes use of post-hoc XAI methods, those which are applied after model training, and prior studies associated with the given datasets and metrics have already evaluated a range of 16 popular methods in the field. The user is provided with a template for each challenge with a generic XAI method function taking a Pytorch [30] model and data as inputs, with explanations as the output.

XAI methods may only be appropriate for certain modalities, i.e. image data rather than natural language data, so it is not necessary that a method has to be submitted to every given challenge. For instance, GradCAM [31] is only usable on Convolutional Neural Network architectures.

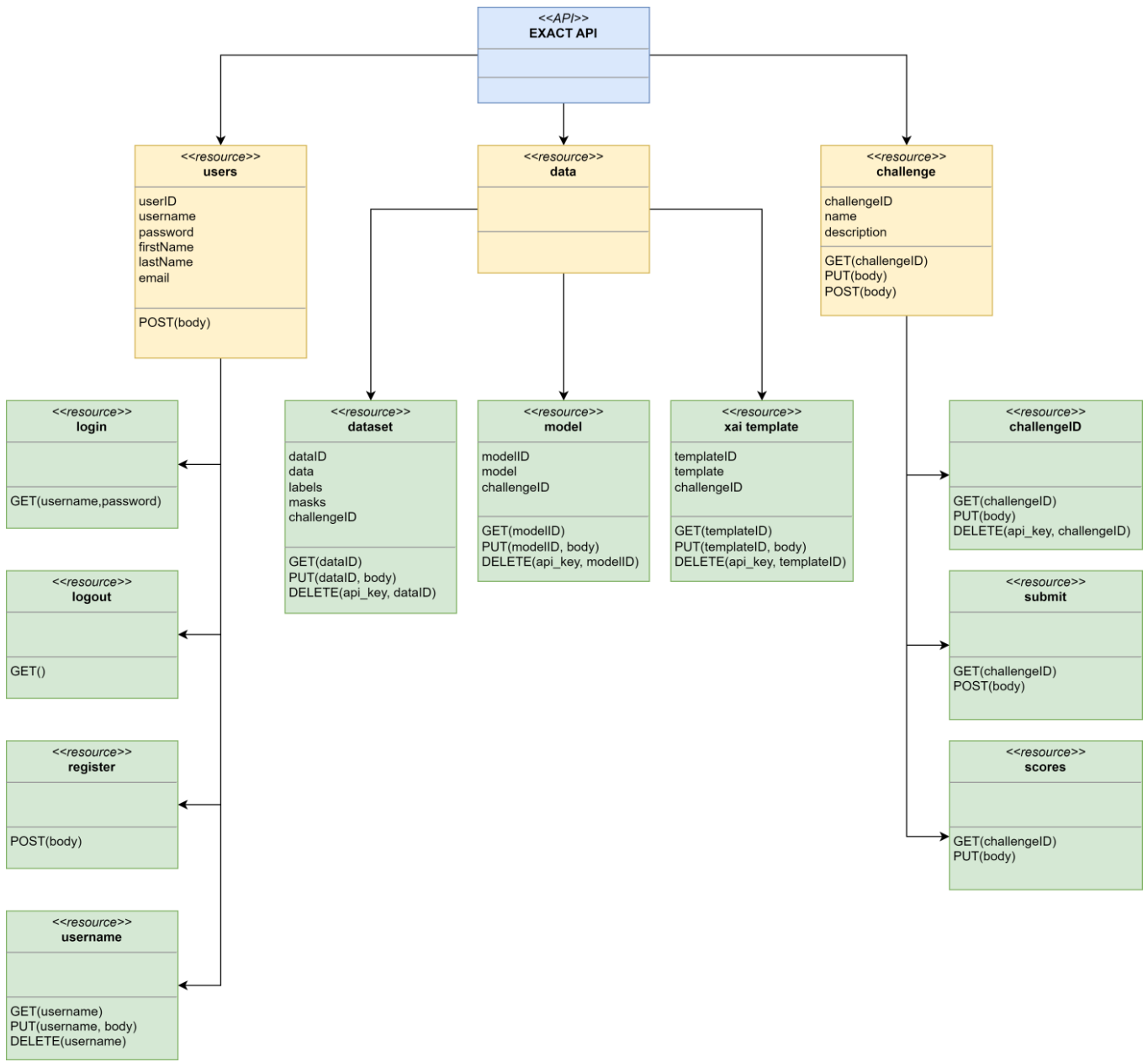

Figure 5. The prototype API diagram for the EXACT platform. Here we can see user, data, and challenge management by the backend service. Standard user management practices are employed for registering new users as well as logging in and out of the platform. Datasets, models, and XAI templates are tied to specific challenges, where the admin uploads these components when creating the challenge. Scores for each challenge are retrieved to be displayed in leaderboards in the frontend. PUT and DELETE commands correspond to creation and deletion of the resource in the relevant database table.

*Evaluation Metrics*

Finally, given the ground truth for explanations and the explanations produced as output by the XAI method, quantitative evaluation metrics are calculated to assess the correctness of explanations. As specified in Section 2.5, each performance metric has its own benefits and characteristics. All metrics relevant to the given challenge are calculated and the resulting performance scores are passed to leaderboards for the challenge. Quantitative performance results for prior work are populated in the leaderboards so that users can see and compare existing results for many popular XAI methods.

*3.2 Technical details and user flow*

Technically, the architecture of the EXACT platform can be broken down into four components – frontend, backend, the database and the worker. Each is composed of separate Docker containers, with the isolated components being able to provide extra security, whilst also being able to communicate with one another as necessary. The frontend is made using Next.js, a React.js framework, serving up the UI to the user. The Django backend connects API routes shown in Figure 5 to frontend components and communicates with a Postgres database to retrieve and store data. The worker is a Python-based container with the libraries required to handle XAI method code, produce explanations, and calculate metric scores.

Once registered and logged in, the user picks a 'challenge' dataset to explain. They are able to download the given data and pre-trained model to locally test their XAI method, as well as the template file to place their XAI code within. Once ready to submit, the user submits the completed template file, where the backend then spawns a worker container with the user's code running. There, explanations are produced for the given XAI method and performance scores are calculated from the array of appropriate metrics for the challenge. Performance scores are then fed back to the backend, which updates the challenge leaderboard for the user to view the results in the frontend.

Deployment at the time of publication will be on publicly available servers hosted by the Physikalisch-Technische Bundesanstalt, the German national institute for metrology, who are well positioned to host and maintain such a platform.

## 4. DISCUSSION AND CONCLUSION

The idea of benchmark platforms is not new to the machine learning community, with existing platforms aiming to benchmark the performance of models across many tasks [32,33]. Such platforms host 'challenges' consisting usually of datasets to be modelled, with users aiming to achieve the best optimised model for the task. In the field of XAI, OpenXAI [26], Quantus [34], and XAI-Bench [35] are three evaluation platforms and toolkits, aiming to assess the performance of XAI methods. OpenXAI provides several tabular ground truth datasets with metrics to evaluate explanations produced, whereas Quantus focuses on the metrics required to evaluate explanations. XAI-Bench provides both ground truth datasets and metrics.

In all cases, metrics used tend to be more 'secondary' criteria, assessing properties of explanations outside of the direct correctness of it. For example, these toolkits use metrics such as robustness and faithfulness, which have been shown to reward explanations that attribute arbitrarily high false-positive attribution to suppressor features [8-11]. More generally, research in [8-11] has shown that multivariate models cannot themselves be understood without knowledge of the underlying data distribution that they were trained on.

Thus, we argue that, while such properties of metrics used in these benchmarks are of interest at a later stage of development in the XAI field, we must first be able to produce consistently correct explanations using data-driven notions of feature importance and 'primary' quantitative metrics.

This is the goal of EXACT – to enable the development of better XAI methods through the use of ground truth benchmark datasets with known feature properties as well as objective and empiric metrics capable of directly assessing explanation performance. Through such a platform, we further the development of reproducible and empirical benchmarks of XAI methods. A future goal would be to unify the benchmarks and metrics present in EXACT with other aforementioned benchmarks, as a key component for strong development of XAI methods is standardised and unified evaluation.

This early prototype succeeds in unifying different modalities of ground truth reference datasets and novel performance metrics to directly quantify explanation performance, with a focus on the issue of false-positive attribution of importance to variables such as suppressor variables. Beyond this, we are building a foundation for future development. Future iterations will focus on refining the processes and user experience of the platform. Such refinements include but are not limited to: new datasets in different modalities (for example, natural language and tabular data), a user management system, improvements to the admin-side challenge creation process, and security features for input file security as well as network security for public hosting purposes.

## 5. CONCLUSION

We have presented EXACT, the Explainable AI Comparison Toolkit, a prototype benchmarking platform for assessing the correctness of explanation methods in the explanations they produce. With a focus on evaluating the false-positive attribution of importance to suppressor features, we incorporate several pre-existing ground truth datasets and novel performance metrics shown to be suited to tackling this problem. We have opened the door for researchers to test their future XAI methods and welcome collaboration to expand the capabilities of EXACT for a better and more standardised future of XAI evaluation.

**ACKNOWLEDGMENTS**

This work is part of a project that has received funding from the European Research Council (ERC) under the European Union's Horizon 2020 research and innovation programme (Grant agreement No. 758985), the German Federal Ministry for Economic Affairs and Climate Action (BMWK) within the "Metrology for Artificial Intelligence in Medicine (M4AIM)" program in the frame of the "QI-Digital'" initiative, and the Heidenhain Foundation in the frame of the Junior Research Group ``Machine Learning and Uncertainty".

**REFERENCES**

1. Rudin C. **Stop Explaining Black Box Machine Learning Models for High Stakes Decisions and Use Interpretable Models Instead**. Nat Mach Intell. 2019;1(5):206–15.

2. Tjoa E, Guan C. **A Survey on Explainable Artificial Intelligence (XAI): Towards Medical XAI**. IEEE Trans Neural Netw Learn Syst. 2020;32(11):4793–813.

3. Doshi-Velez F, Kortz M, Budish R, Bavitz C, Gershman S, O'Brien D, et al. **Accountability of AI Under the Law: The Role of Explanation**. ArXiv171101134 Cs Stat [Internet]. 2019 Dec 20 [cited 2020 Oct 19]; Available from: http://arxiv.org/abs/1711.01134

4. Lundberg SM, Lee SI. **A Unified Approach to Interpreting Model Predictions**. In: Guyon I, Luxburg UV, Bengio S, Wallach H, Fergus R, Vishwanathan S, et al., editors. Advances in Neural Information Processing Systems 30 [Internet]. Curran Associates, Inc.; 2017 [cited 2020 Oct 7]. p. 4765–74. Available from: http://papers.nips.cc/paper/7062-a-unified-approach-to-interpreting-model-predictions.pdf

5. Montavon G, Lapuschkin S, Binder A, Samek W, Müller KR. **Explaining nonlinear classification decisions with deep taylor decomposition**. Pattern Recognit. 2017;65:211–22.

6. Ribeiro MT, Singh S, Guestrin C. **Why should i trust you?: Explaining the predictions of any classifier**. In: Proceedings of the 22nd ACM SIGKDD international conference on knowledge discovery and data mining. ACM; 2016. p. 1135–44.

7. Haufe S, Meinecke F, Görgen K, Dähne S, Haynes JD, Blankertz B, et al. **On the interpretation of weight vectors of linear models in multivariate neuroimaging**. NeuroImage. 2014 Feb 15;87:96–110.

8. Wilming R, Kieslich L, Clark B, Haufe S. **Theoretical Behavior of XAI Methods in the Presence of Suppressor Variables**. In: Krause A, Brunskill E, Cho K, Engelhardt B, Sabato S, Scarlett J, editors. Proceedings of the 40th International Conference on Machine Learning [Internet]. PMLR; 2023. p. 37091–107. (Proceedings of Machine Learning Research; vol. 202). Available from: https://proceedings.mlr.press/v202/wilming23a.html

9. Wilming R, Budding C, Müller KR, Haufe S. **Scrutinizing XAI using linear ground-truth data with suppressor variables**. Mach Learn. 2022 May 1;111(5):1903–23.

10. Clark B, Wilming R, Haufe S. **XAI-TRIS: Non-linear benchmarks to quantify ML explanation performance** [**Internet**]. arXiv; 2023 [cited 2023 Aug 29]. Available from: http://arxiv.org/abs/2306.12816

11. Oliveira M, Wilming R, Clark B, Budding C, Eitel F, Ritter K, et al. **Benchmarking the influence of pre-training on explanation performance in MR image classification**. Front Artif Intell [Internet]. 2024;7. Available from: https://www.frontiersin.org/articles/10.3389/frai.2024.1330919

12. Kokhlikyan N, Miglani V, Martin M, Wang E, Alsallakh B, Reynolds J, et al. **Captum: A unified and generic model interpretability library for PyTorch**. ArXiv200907896 Cs Stat [Internet]. 2020 Sep 16 [cited 2020 Nov 30]; Available from: http://arxiv.org/abs/2009.07896

13. Alber M, Lapuschkin S, Seegerer P, Hägele M, Schütt KT, Montavon G, et al. **iNNvestigate neural networks!** ArXiv180804260 Cs Stat [Internet]. 2018 Aug 13 [cited 2020 Oct 19]; Available from: http://arxiv.org/abs/1808.04260

14. Golomb SW. **Polyominoes: puzzles, patterns, problems, and packings**. Vol. 111. Princeton University Press; 1996.

15. Nintendo of America. **Tetris** [**Internet**]. Nintendo Entertainment System. Redmond, WA : Nintendo of America, [1989] ©1989; 1989. Available from: https://search.library.wisc.edu/catalog/9910796303302121

16. Deng J, Dong W, Socher R, Li L, Kai Li, Li Fei-Fei. **ImageNet: A Large-Scale Hierarchical Image Database**. In: 2009 IEEE Conference on Computer Vision and Pattern Recognition. 2009. p. 248–55.

17. Conger AJ. **A Revised Definition for Suppressor Variables: a Guide To Their Identification and Interpretation , A Revised Definition for Suppressor Variables: a Guide To Their Identification and Interpretation**. Educ Psychol Meas. 1974 Apr 1;34(1):35–46.

18. Friedman L, Wall M. **Graphical Views of Suppression and Multicollinearity in Multiple Linear Regression**. Am Stat. 2005 May 1;59(2):127–36.

19. Van Essen DC, Smith SM, Barch DM, Behrens TEJ, Yacoub E, Ugurbil K. **The WU-Minn Human Connectome Project: An overview**. NeuroImage. 2013 Oct 15;80:62–79.

20. Jenkinson M, Beckmann CF, Behrens TEJ, Woolrich MW, Smith SM. **FSL**. NeuroImage. 2012;62(2):782–90.

21. Fischl B. **FreeSurfer**. NeuroImage. 2012 Aug 15;62(2):774–81.

22. Milchenko M, Marcus D. **Obscuring Surface Anatomy in Volumetric Imaging Data**. Neuroinformatics. 2013 Jan 1;11(1):65–75.

23. Otsu N. **A Threshold Selection Method from Gray-Level Histograms**. IEEE Trans Syst Man Cybern. 1979;9(1):62–6.

24. Wharton SB, Simpson JE, Brayne C, Ince PG. **Age-Associated White Matter Lesions: The MRC Cognitive Function and Ageing Study**. Brain Pathol. 2015;25(1):35–43.

25. d'Arbeloff T, Elliott ML, Knodt AR, Melzer TR, Keenan R, Ireland D, et al. **White matter hyperintensities are common in midlife and already associated with cognitive decline**. Brain Commun. 2019 Dec;1(1).

26. Agarwal C, Krishna S, Saxena E, Pawelczyk M, Johnson N, Puri I, et al. **OpenXAI: Towards a transparent evaluation of model explanations**. Adv Neural Inf Process Syst. 2022;35:15784–99.

27. Bonneel N, Van De Panne M, Paris S, Heidrich W. **Displacement interpolation using Lagrangian mass transport**. In: Proceedings of the 2011 SIGGRAPH Asia conference. 2011. p. 1–12.

28. Flamary R, Courty N, Gramfort A, Alaya MZ, Boisbunon A, Chambon S, et al. **POT: Python Optimal Transport**. J Mach Learn Res. 2021;22(78):1–8.

29. Arras L, Osman A, Samek W. **CLEVR-XAI: A benchmark dataset for the ground truth evaluation of neural network explanations**. Inf Fusion. 2022;81:14–40.

30. Paszke A, Gross S, Massa F, Lerer A, Bradbury J, Chanan G, et al. **PyTorch: An Imperative Style, High-Performance Deep Learning Library**. In: Advances in Neural Information Processing Systems 32 [Internet]. Curran Associates, Inc.; 2019. p. 8024–35. Available from: http://papers.neurips.cc/paper/9015-pytorch-an-imperative-style-high-performance-deep-learning-library.pdf

31. Selvaraju RR, Cogswell M, Das A, Vedantam R, Parikh D, Batra D. **Grad-CAM: Visual Explanations from Deep Networks via Gradient-Based Localization**. In: 2017 IEEE International Conference on Computer Vision (ICCV). 2017. p. 618–26.

32. Vanschoren J, van Rijn JN, Bischl B, Torgo L. **OpenML: networked science in machine learning**. ACM SIGKDD Explor Newsl. 2014 Jun 16;15(2):49–60.

33. Yadav D, Jain R, Agrawal H, Chattopadhyay P, Singh T, Jain A, et al. **EvalAI: Towards Better Evaluation Systems for AI Agents**. CoRR [Internet]. 2019;abs/1902.03570. Available from: http://arxiv.org/abs/1902.03570

34. Hedström A, Weber L, Bareeva D, Motzkus F, Samek W, Lapuschkin S, et al. **Quantus: An Explainable AI Toolkit for Responsible Evaluation of Neural Network Explanations** [**Internet**]. arXiv; 2022. Available from: https://arxiv.org/abs/2202.06861

35. Liu Y, Khandagale S, White C, Neiswanger W. **Synthetic Benchmarks for Scientific Research in Explainable Machine Learning**. CoRR [Internet]. 2021;abs/2106.12543. Available from: https://arxiv.org/abs/2106.12543